\documentclass[11pt,a4paper]{article}

\usepackage[T1]{fontenc}
\usepackage[margin=1in]{geometry}
\usepackage{amsmath}
\usepackage{amssymb}
\usepackage[nopatch]{microtype}
\usepackage{booktabs}
\usepackage{graphicx}
\usepackage{multirow}
\usepackage{algorithm}
\usepackage{algpseudocode}
\usepackage[dvipsnames,table]{xcolor}
\usepackage{threeparttable}
\usepackage{hyperref}

\usepackage[backend=biber,style=authoryear,maxcitenames=2,maxbibnames=99,doi=false,url=true,isbn=false]{biblatex}
\newcommand{\headrow}{\rowcolor{gray!15}}

\title{Cross-Family Speculative Decoding for Polish Language Models on Apple~Silicon:\\
An Empirical Evaluation of Bielik~11B with UAG-Extended MLX-LM}

\author{Krzysztof Fona\l{}\\
\small Wroc\l{}aw University of Science and Technology, Poland\\
\small \texttt{krzysztof.fonal@pwr.edu.pl}}

\newcommand{\keywords}[1]{\par\noindent\textbf{Keywords:} #1}

\begin{document}

\maketitle

\begin{abstract}
Speculative decoding accelerates large language model inference by using a smaller \emph{draft} model to propose $k$ candidate tokens that a larger \emph{target} model verifies in a single forward pass, amortising the cost of weight loading across multiple candidate positions. While effective for same-tokenizer model pairs on high-bandwidth GPU hardware, its applicability to cross-family pairs with mismatched tokenizers, to morphologically rich inflected languages, and to consumer-grade unified memory architectures remains underexplored. In this work, we extend the MLX-LM inference framework with Universal Assisted Generation (UAG), enabling cross-tokenizer speculative decoding on Apple Silicon. We evaluate Bielik~11B-Instruct --- a Polish language model based on the Mistral architecture --- as the target model, paired with three draft models from distinct model families: Bielik~1.5B (Qwen2.5-based with a custom Polish tokenizer), Qwen2.5-1.5B, and Llama~3.2-1B. Experiments are conducted across three Polish-language datasets (Polish Wikipedia, pl\_alpaca, and synthetically generated short questions) with draft lengths $k \in \{2, 4\}$, plus a targeted confirmatory experiment at $k = 6$. We evaluate both naive and context-aware token translation strategies for cross-tokenizer alignment. Our results reveal that: (1)~context-aware translation consistently improves token acceptance rates over both naive and no-translation conditions across all drafter--dataset combinations; (2)~the Polish-specialised Bielik~1.5B drafter, despite targeting the same language as the verifier, achieves lower acceptance rates than the general-purpose Qwen2.5-1.5B and Llama~3.2-1B drafters; (3)~speculative decoding throughput on Apple~Silicon is strongly content-dependent, achieving up to $1.7\times$ speedup for structured or repetitive text while failing to outperform autoregressive decoding for varied instruction-following content; and (4)~verification cost on unified memory hardware does not amortise across $k$ positions as theory predicts, because both draft and target model inference are memory-bandwidth bound, making $k$ sequential draft passes prohibitively expensive relative to the benefit gained from batched verification. We propose a hardware-aware parametric speedup formula grounded in these observations and characterise the conditions under which cross-family speculative decoding can benefit inflected-language generation on Apple Silicon. To our knowledge, this is the first systematic evaluation of cross-family speculative decoding for Polish language models and the first empirical study of UAG-based speculative decoding on Apple Silicon's unified memory architecture.
\end{abstract}


%% ============================================================
\section{Introduction}
\label{sec:introduction}
%% ============================================================
 
The public release of ChatGPT \autocite{openai2022chatgpt} in late 2022 marked a turning point in the practical accessibility of large language model (LLM) capabilities, catalysing rapid adoption across scientific, professional, and consumer domains. Beyond static question answering, LLMs increasingly underpin autonomous \emph{agentic} systems that operate in extended iterative loops \autocite{wang2024agents}: coding assistants such as Cursor\footnote{\url{https://www.cursor.com}} and GitHub Copilot \autocite{chen2021codex} accelerate software development by generating, editing, and testing code in tightly integrated feedback cycles, while general-purpose personal agents interact directly with file systems, messaging platforms, and external services to complete multi-step tasks on behalf of users. The privacy implications of this expanding scope are acute. OpenClaw \autocite{openclaw2024} --- an open-source personal AI assistant that has accumulated over 300,000 GitHub stars --- exemplifies the concern: the platform assists users with personal tasks by drawing local files, emails, and other private data into the model's context, using messaging applications such as WhatsApp, Telegram, and iMessage as a convenient conversational interface. When backed by a cloud-hosted language model, every document, email thread, and personal file that enters the context window is transmitted to a third-party server. For individual users, journalists, legal professionals, and enterprises handling confidential information, this represents an unacceptable exposure. Privacy is not the only pressure, however. As LLM usage grows, so do API costs, and the dominant cloud providers currently operate their flagship models at a loss --- a pricing structure that analysts expect to be corrected over time as competitive pressures ease. Organisations that build workflows around cloud-hosted inference therefore carry an implicit cost and availability risk tied to the pricing decisions of a handful of vendors. Together, these two concerns --- data sovereignty and long-term cost predictability --- are primary drivers of growing interest in \emph{locally deployed} LLMs, where all inference runs on the user's own hardware, no data leaves the device, and the marginal cost of each query is bounded by the user's own electricity bill.
 
Apple Silicon hardware has emerged as a compelling and cost-accessible platform for local LLM inference. The M-series system-on-chip designs integrate CPU, GPU, and Neural Engine on a single die with a shared unified memory pool, eliminating the PCIe data transfer bottleneck of discrete GPU systems. From a cost and power perspective, a 14-inch MacBook Pro equipped with the M2 Pro (10 CPU cores, 16 GPU cores, 32~GB unified memory, 200~GB/s memory bandwidth) retails for approximately one tenth the cost of a complete single-GPU NVIDIA A100 server node --- the card alone exceeds \$10{,}000, with chassis, cooling, and power infrastructure adding substantially to the total --- 
while consuming an order of magnitude less power under inference load: the A100 PCIe card alone carries a 300~W thermal design rating \autocite{nvidia_a100_datasheet}, against approximately 30~W for the M2 Pro MacBook Pro system \autocite{apple_m2pro_spec}, before accounting for additional server cooling overhead. The MLX framework \autocite{mlx2023} and its LLM-specific extension MLX-LM \autocite{mlxlm2024} provide a mature inference stack for these devices, making Apple Silicon a practically accessible deployment target for researchers and developers running models locally.
 
Achieving practical inference speeds on consumer hardware is an active area of systems research, spanning model compression, kernel optimisation, batching strategies, and architectural changes to the decoding loop itself. Among the earliest techniques to achieve widespread adoption, \emph{quantization} \autocite{dettmers2022llmint8} reduces the numerical precision of model weights from 32- or 16-bit floating point to 8- or 4-bit integers, compressing model size two- to four-fold with modest quality degradation, and is now standard practice for consumer deployment. A complementary line of work targets the sequential bottleneck of autoregressive generation: \emph{speculative decoding} \autocite{leviathan2023fast,chen2023accelerating} uses a smaller, faster \emph{draft} model to propose $k$ candidate tokens that a larger \emph{target} model verifies in a single forward pass, without altering the output distribution. The central assumption is that on memory-bandwidth-bound hardware, verifying $k$ candidates costs approximately the same as generating a single token autoregressively, since the dominant cost --- loading model weights from memory --- is incurred once regardless of the number of candidates \autocite{xia2024decoding}. When draft proposals are accurate, this yields near-$k\times$ throughput improvement at no loss in generation quality.
 
Most existing speculative decoding work, however, assumes draft and target models share the same tokenizer --- a constraint that restricts draft model selection to the same model family. For widely spoken languages supported by large research communities, this is rarely a barrier: major families such as Llama \autocite{llama32_2024} and Qwen \autocite{qwen25_2024} each span parameter scales from 1B to 70B\texttt{+} with a shared tokenizer. For under-resourced languages, the situation is more constrained. Polish, like other morphologically rich Slavic languages, requires LLMs specifically adapted to its inflectional morphology, and the ecosystem of Polish-language models is currently dominated by the Bielik family \autocite{bielik11b2025}. Crucially, the Bielik family is not tokenizer-homogeneous: Bielik~11B~v3, the primary large Polish model, is based on the Mistral architecture with a Mistral-derived tokenizer ($\sim$32K vocabulary entries), while Bielik~1.5B~v3 is adapted from Qwen2.5 and uses the APT4 morphological tokenizer --- a custom vocabulary designed for Polish orthography and inflectional morphology. Despite belonging to the same Polish LLM family, the two models are tokenizer-incompatible, ruling out standard same-tokenizer speculative decoding.
 
Universal Assisted Generation (UAG) \autocite{uag2024} addresses this limitation by introducing string-level token translation between vocabularies, enabling cross-family draft--target pairs in which the draft and target tokenizers may differ arbitrarily. A context-aware refinement improves alignment by using surrounding token context to resolve tokenisation ambiguities at vocabulary boundaries. While UAG has since been extended with a Token-Level Intersection approach \autocite{uag_tli2024} and integrated into the HuggingFace Transformers library, neither variant has been ported to Apple's MLX framework, leaving Apple Silicon users without access to cross-tokenizer speculative decoding.
 
Two gaps therefore remain unaddressed in the literature. First, UAG-based cross-tokenizer speculative decoding is unavailable in MLX-LM, preventing practitioners on Apple Silicon from exploiting cross-family draft models. Second, no systematic empirical study of speculative decoding exists for Polish language models, nor for unified memory hardware under realistic inference conditions. In particular, it is unknown whether the bandwidth-bound cost model that underpins speculative decoding's theoretical advantage holds on Apple Silicon's unified memory architecture, where both draft and target model inference compete for the same memory bus. This paper addresses both gaps. We extend MLX-LM with full UAG support and conduct the first systematic evaluation of cross-family speculative decoding for Polish language models on Apple Silicon, characterising when and why the approach succeeds or fails.
 
We make the following contributions:
 
\begin{enumerate}
    \item We implement UAG in MLX-LM, including both naive token translation and a context-aware strategy that uses a configurable token-prefix window to resolve vocabulary boundary ambiguities, enabling cross-family speculative decoding for any model pair supported by the framework.
    \item We conduct the first systematic evaluation of Bielik~11B-Instruct as a speculative decoding target with three cross-family draft models --- Bielik~1.5B, Qwen2.5-1.5B, and Llama~3.2-1B --- across three Polish-language datasets, finding that general-purpose drafters unexpectedly outperform the Polish-specialised drafter due to tokenizer incompatibility introduced by the APT4 morphological vocabulary.
    \item We demonstrate empirically that verification cost on unified memory hardware does not amortise across $k$ positions as standard theory predicts, and derive a hardware-aware parametric speedup formula that accurately captures our observed break-even thresholds and predicts performance on higher-bandwidth Apple Silicon hardware.
    \item We characterise the content-type conditions and draft length constraints under which cross-family speculative decoding achieves practical speedups on Apple Silicon, providing actionable deployment guidelines for practitioners working with inflected-language models on consumer hardware.
\end{enumerate}
 
The remainder of this paper is organised as follows. Section~\ref{sec:related} surveys related work on speculative decoding, cross-tokenizer methods, Polish language models, and Apple Silicon inference. Section~\ref{sec:method} describes our UAG integration into MLX-LM and the token translation strategies employed. Section~\ref{sec:setup} details the experimental setup, including models, datasets, hardware, and evaluation metrics. Section~\ref{sec:results} presents the experimental results. Section~\ref{sec:analysis} provides an in-depth analysis of the performance characteristics observed on unified memory hardware. Section~\ref{sec:conclusion} concludes with practical recommendations and directions for future work.

%% ============================================================
\section{Related Work}
\label{sec:related}
%% ============================================================
 
\subsection{Speculative Decoding}
 
An early precursor to speculative decoding is blockwise parallel decoding \autocite{stern2018blockwise}, which proposed attaching multiple prediction heads to the target model to predict several future tokens simultaneously, accepting the longest correct prefix. Speculative decoding in its modern form was independently proposed by Leviathan et~al.\ \autocite{leviathan2023fast} and Chen et~al.\ \autocite{chen2023accelerating}, who demonstrated that a separate, smaller draft model can propose candidate tokens for verification by a larger target model without altering the target's output distribution. The key insight is that verifying $k$ candidate tokens in a single forward pass is substantially cheaper than generating them autoregressively when inference is memory-bandwidth-bound, as the model weights need to be loaded from memory only once regardless of the number of candidates. A comprehensive analysis of speculative decoding's practical performance characteristics is provided by \textcite{xia2024decoding}.
 
Subsequent work has extended the paradigm in several directions. Medusa \autocite{cai2024medusa} eliminates the separate draft model entirely by attaching multiple lightweight draft heads directly to the target model, enabling tree-structured verification of multiple candidate sequences in a single forward pass. EAGLE \autocite{li2024eagle} takes a complementary approach, training a lightweight autoregressive model that operates on the target model's hidden feature representations rather than the token space, achieving higher acceptance rates at comparable draft model sizes. ReDrafter \autocite{zhang2024redrafter}, developed by Apple, uses a recurrent architecture for the draft model to minimise per-step latency on resource-constrained hardware, reporting approximately 2.3$\times$ end-to-end speedup. More recent work explores nested speculation strategies \autocite{kumar2026speculative}. All of these approaches assume a shared tokenizer between draft and target; the cross-tokenizer case is addressed in Section~\ref{subsec:cross_tokenizer}.

\subsection{Cross-Tokenizer Speculative Decoding}
\label{subsec:cross_tokenizer}
 
Standard speculative decoding requires the draft and target models to share the same tokenizer. Several recent works have addressed this limitation:
 
\textbf{Universal Assisted Generation (UAG).} Developed by Intel Labs and HuggingFace, UAG addresses cross-tokenizer speculative decoding through two successive approaches. The first \autocite{uag2024} introduces the SLEM (String-Level Exact Matching) algorithm \autocite{timor2025acceleratingllminferencelossless}: draft tokens are decoded to a surface string, re-tokenized in the target vocabulary, and the inverse mapping is applied after verification — a training-free round-trip that works with any model pair. The same post identifies that accurate retokenization requires prepending a short window of previously accepted tokens as context, without which boundary misalignment sharply degrades acceptance rates. The second approach \autocite{uag_tli2024}, following the Token-Level Intersection (TLI) method of \textcite{timor2025acceleratingllminferencelossless}, replaces the string round-trip entirely: at sampling time, the draft model is constrained to draw only from the \emph{intersection} of the draft and target vocabularies, eliminating translation latency but restricting the draft to tokens present in both vocabularies simultaneously. The HuggingFace Transformers library currently implements TLI as its primary UAG strategy. In this work we implement and evaluate the SLEM variant with a context-aware prefix window for MLX-LM, rather than TLI, because the intersection of the Mistral-based Bielik~11B tokenizer with the Qwen and Llama tokenizers is too small to support fluent Polish generation --- the majority of Polish-specific tokens, diacritics, and morphological suffixes are absent from the intersection, which would severely degrade draft model fluency and, consequently, acceptance rates.
 
\textbf{OmniDraft.} Presented at NeurIPS 2025, OmniDraft \autocite{omnidraft2025} introduces a cross-vocabulary online adaptive drafter that uses an n-gram cache to translate between vocabularies, enabling a single Llama-68M model to pair with target models from different families. OmniDraft is complementary rather than directly comparable to our approach: it requires training a dedicated universal draft model, whereas UAG-based translation works with any existing model pair without additional training. For Polish, no pre-trained OmniDraft-style universal drafter exists, making training-free cross-tokenizer methods such as UAG the only currently practical option.
 
% TODO: Add TokenTiming, Cross-Attention Speculative Decoding if relevant

\subsection{Polish Language Models}
 
The Bielik family of models represents the primary open-source effort for Polish-optimized LLMs. Importantly for this study, the Bielik family spans multiple base architectures:
 
\textbf{Bielik~11B~v3} \autocite{bielik11b2025} is based on the Mistral~7B~v0.2 architecture, scaled to 11B parameters via depth up-scaling from 32 to 50 layers. It uses a Mistral-derived tokenizer with approximately 32,128 vocabulary entries.
 
\textbf{Bielik~v3~Small} \autocite{bieliksmall2025} (1.5B and 4.5B) are adapted from Qwen2.5 base models (1.5B and 3B respectively) with depth up-scaling, a replaced custom Polish tokenizer (APT4), and continued pretraining on 292 billion tokens.
 
This architectural divergence within the Bielik family has an important implication: even Bielik 1.5B is not a ``same-family'' draft model for Bielik 11B in the speculative decoding sense, as they differ in both architecture (Qwen2.5 vs.\ Mistral) and tokenizer (APT4 vs.\ Mistral).

\subsection{LLM Inference on Apple Silicon}
 
Apple's MLX framework \autocite{mlx2023} provides a NumPy-like API for machine learning on Apple Silicon, with operations compiled to Metal shaders for GPU execution. The MLX-LM library \autocite{mlxlm2024} extends MLX with LLM-specific optimisations including KV caching and speculative decoding support for same-tokenizer pairs. \textcite{barrios2025native} provide a systematic benchmarking study of LLM inference on Apple Silicon across model sizes and quantisation levels. The unified memory architecture eliminates CPU--GPU data transfer overhead, but the shared memory bus remains the binding constraint: draft and target model inferences run sequentially, and each forward pass must stream its full weight set from DRAM to the GPU compute units, making memory bandwidth the dominant cost for both.
 
A central hardware context for our study is the bandwidth gap between Apple Silicon and data-centre accelerators. The M2 Pro's 200~GB/s unified memory bandwidth is approximately one tenth of the NVIDIA A100's 2~TB/s HBM2e bandwidth \autocite{nvidia_a100_datasheet,apple_m2pro_spec}. This gap matters for speculative decoding because the technique's theoretical speedup assumes that verification of $k$ tokens costs approximately as much as generating one token autoregressively — an assumption grounded in the memory-bandwidth-bound nature of inference. Whether this assumption holds under Apple Silicon's unified memory model, where CPU and GPU share the same bus, is an open empirical question that this paper addresses.
 
Hardware-aware speculative decoding for Apple Silicon has seen limited prior work. The closest prior art is ReDrafter \autocite{zhang2024redrafter}, Apple's own recurrent-draft system designed for on-device inference, which reports approximately 2.3$\times$ speedup on same-tokenizer model pairs. Our work differs in two key respects: we target cross-tokenizer pairs via UAG-style token translation, and we provide a systematic empirical characterisation of when and why the bandwidth-bound cost model breaks down on unified memory hardware.

%% ============================================================
\section{Method}
\label{sec:method}
%% ============================================================
 
\subsection{MLX-LM Speculative Decoding Baseline}
 
The MLX-LM library provides a speculative decoding implementation that supports same-tokenizer draft--target pairs. At each decoding step, the draft model generates $k$ candidate tokens autoregressively, and the target model verifies all $k$ candidates in a single forward pass using causal attention masking. Tokens are accepted sequentially until the first rejection; the target model's prediction at the rejection point replaces the rejected token, and the process repeats.
 
\subsection{Universal Assisted Generation in MLX-LM}
 
We extend MLX-LM's speculative decoding pipeline \autocite{mlxlm2024} with UAG support \autocite{uag2024}, enabling draft and target models with different tokenizers. Our implementation adds a token translation layer between the draft generation and verification stages.
 
\subsubsection{Naive Token Translation}
 
The naive translation strategy converts draft tokens to the target vocabulary purely via string round-trip, without any surrounding context --- equivalent to the SLEM algorithm of \textcite{timor2025acceleratingllminferencelossless}. Algorithm~\ref{alg:naive} gives the full decoding loop.
 
\begin{algorithm}[t]
\caption{UAG Decoding with Naive Token Translation}
\label{alg:naive}
\begin{algorithmic}[1]
\Require Draft model $M_d$ with tokenizer $\mathcal{T}_d$; target model $M_t$ with tokenizer $\mathcal{T}_t$; accepted context $c$ (in $\mathcal{T}_d$ token ids); number of draft tokens $k$
\Ensure Updated context $c$ extended by one or more accepted tokens
\While{generation not complete}
    \State $\mathbf{d} \gets M_d.\text{generate}(c,\; k)$ \Comment{$k$ draft tokens in $\mathcal{T}_d$}
    \State $s \gets \mathcal{T}_d.\text{decode}(\mathbf{d})$ \Comment{draft tokens $\to$ text}
    \State $\mathbf{v} \gets \mathcal{T}_t.\text{encode}(s)$ \Comment{text $\to$ target token ids}
    \State $\mathbf{a},\, t^* \gets M_t.\text{verify}(c_t,\; \mathbf{v})$ \Comment{accept mask + bonus token}
    \State $j \gets $ first index where $\mathbf{a}[j] = 0$, or $|\mathbf{v}|$ if all accepted
    \State $s' \gets \mathcal{T}_t.\text{decode}(\mathbf{v}_{1:j})$ \Comment{accepted target tokens $\to$ text}
    \State $c \gets c \,\|\, \mathcal{T}_d.\text{encode}(s')$ \Comment{back-translate to draft vocab}
\EndWhile
\end{algorithmic}
\end{algorithm}
 
This approach introduces a fundamental challenge: token boundaries in the draft and target tokenizers may not align. A single draft token may correspond to multiple target tokens, or vice versa, depending on how each tokenizer segments the same character sequence. Critically, modern BPE-based tokenizers assign different token identifiers to the same surface form depending on its position and surrounding context --- in particular, whether a character or substring appears at the start of a sequence, at a word boundary after a space, or inside a word. This is especially consequential for Polish, where single-character function words are common. Consider the letter \emph{w}: as a standalone token it is the Polish preposition meaning \emph{in} (e.g.\ \emph{w Warszawie}, `in Warsaw'), whereas the same character also appears word-initially in nouns such as \emph{wino} (`wine'), which a tokenizer may segment as \emph{w\,+\,ino}. Because the positional context differs, the two surface occurrences of \emph{w} receive distinct token identifiers in the vocabulary. Naive string-level round-trip translation discards this positional information, so the re-tokenized candidate sequence may not match what the target tokenizer would produce in context --- causing spurious rejections even when the draft and target models agree on the underlying text.
 
\subsubsection{Context-Aware Token Translation}
 
To mitigate boundary misalignment, we develop a context-aware string-level translation strategy motivated by the observation in \textcite{uag2024} that accurate re-tokenization requires prepending a window of preceding accepted tokens. The original UAG blog post identifies this requirement but does not present a concrete algorithm. The HuggingFace Transformers implementation subsequently evolved to incorporate Token-Level Intersection (TLI), introduced by \textcite{uag_tli2024} and formally analysed by \textcite{timor2025acceleratingllminferencelossless}: rather than translating through text, TLI constrains the draft model at sampling time to draw only from the \emph{intersection} of the draft and target vocabularies, eliminating the string round-trip overhead of SLEM entirely. However, TLI introduces a hard coverage constraint --- any token absent from the intersection is permanently excluded from draft proposals --- which is particularly damaging for the language pairs in this study. Pairing Bielik~11B (Mistral tokenizer, $\sim$32K vocabulary) with general-purpose drafters such as Qwen2.5 or Llama~3.2 ($\sim$128K--152K vocabularies) yields an intersection dominated by common Latin characters, digits, and punctuation; the majority of Polish-specific tokens, diacritics, and morphological suffixes are unique to one side. Operating within such a restricted vocabulary would substantially degrade draft model fluency in Polish and, consequently, the token acceptance rate. TLI for MLX-LM is therefore left as a direction for future work. Neither the string-level nor the TLI approach is currently available in MLX-LM, as both HuggingFace implementations are written in PyTorch and tightly coupled to that framework's KV cache management, input tensor conventions, and speculative decoding loop --- porting them to MLX is non-trivial given its distinct execution model (lazy evaluation via Metal shaders), different cache synchronisation semantics, and different conventions for returning accepted tokens to the draft model.
 
Our context-aware translation for MLX-LM is therefore an independent implementation. The core idea is to prepend a prefix of $p$ previously accepted tokens before re-tokenization, then strip those prefix tokens after encoding so that only the genuinely new draft candidates are passed to the verifier, giving the tokenizer sufficient positional context to correctly segment the boundary. Algorithm~\ref{alg:ctx_aware} presents the high-level decoding loop; engineering details specific to the MLX framework are discussed in Section~\ref{subsec:impl_details}.
 
\begin{algorithm}[t]
\caption{UAG Decoding with Context-Aware Token Translation}
\label{alg:ctx_aware}
\begin{algorithmic}[1]
\Require As in Algorithm~\ref{alg:naive}; additionally context-prefix length $p$
\Ensure Updated context $c$ extended by one or more accepted tokens
\While{generation not complete}
    \State $\mathbf{d} \gets M_d.\text{generate}(c,\; k)$
    \State $s_\text{prefix} \gets \mathcal{T}_d.\text{decode}(c_{-p:})$ \Comment{\textbf{last $p$ accepted tokens as text}}
    \State $s \gets \mathcal{T}_d.\text{decode}(\mathbf{d})$
    \State $\mathbf{v}_\text{full} \gets \mathcal{T}_t.\text{encode}(s_\text{prefix} \,\|\, s)$ \Comment{\textbf{encode prefix + draft together}}
    \State $\ell \gets |\mathcal{T}_t.\text{encode}(s_\text{prefix})|$ \Comment{\textbf{prefix length in target tokens}}
    \State $\mathbf{v} \gets \mathbf{v}_\text{full}[\ell:]$ \Comment{\textbf{strip prefix; keep only draft candidates}}
    \State $\mathbf{a},\, t^* \gets M_t.\text{verify}(c_t,\; \mathbf{v})$
    \State $j \gets $ first index where $\mathbf{a}[j] = 0$, or $|\mathbf{v}|$ if all accepted
    \State $s' \gets \mathcal{T}_t.\text{decode}(\mathbf{v}_{1:j})$
    \State $c \gets c \,\|\, \mathcal{T}_d.\text{encode}(s')$
\EndWhile
\end{algorithmic}
\end{algorithm}

\subsection{Implementation Details}
\label{subsec:impl_details}
 
Our implementation is publicly available as a fork of MLX-LM.\footnote{MLX-LM fork: \url{https://github.com/krzysiekfonal/mlx-lm/tree/feature/slem-with-context-aware}.} The core modification is to the \texttt{speculative\_generate\_step} function in \texttt{generate.py}, which is the central routine responsible for one full draft--verify cycle in MLX-LM's speculative decoding pipeline. We extend its signature with two new parameters: \texttt{draft\_tokenizer}, which accepts an optional second tokenizer instance for the draft model, and \texttt{translation\_prefix\_tokens}, which controls the context window size $p$ used during context-aware re-tokenization. When \texttt{draft\_tokenizer} is \texttt{None} the function behaves identically to the original same-tokenizer implementation, preserving backward compatibility.
 
When a draft tokenizer is provided, the modified function performs the full token translation loop described in Algorithms~\ref{alg:naive} and~\ref{alg:ctx_aware}. Beyond the translation steps shown in the pseudocode, the implementation handles three additional concerns specific to the MLX framework: (1)~bidirectional vocabulary conversion between draft and target token id spaces; (2)~KV cache synchronisation after each verification step, since the accepted sequence length in target tokens may differ from the corresponding draft token count after translation; and (3)~back-translation of accepted target tokens into the draft tokenizer's vocabulary before updating the draft model's context for the next iteration.

%% ============================================================
\section{Experimental Setup}
\label{sec:setup}
%% ============================================================

\subsection{Models}

Table~\ref{tab:models} summarizes the models used in our experiments. All models are evaluated in their MLX-converted form to ensure consistent execution through the same Metal GPU backend. The target model is quantized to 8-bit integers and the three draft models to 4-bit integers, following standard practice for consumer-hardware deployment \autocite{dettmers2022llmint8}. Bielik~11B \autocite{bielik11b2025} serves as the target; Bielik~1.5B \autocite{bieliksmall2025}, Qwen2.5-1.5B \autocite{qwen25_2024}, and Llama~3.2-1B \autocite{llama32_2024} serve as draft models.

\begin{table}[hbt!]
\begin{threeparttable}
\caption{Models used in experiments. All models are cross-family relative to the Bielik~11B target.}
\label{tab:models}
\begin{tabular}{llll}
\toprule
\headrow Model & Role & Base Arch. & Tokenizer \\
\midrule
Bielik 11B v3 Instruct & Target & Mistral & Mistral ($\sim$32K) \\
Bielik 1.5B v3 Instruct & Draft & Qwen2.5 & APT4 (Polish) \\
Qwen2.5 1.5B Instruct & Draft & Qwen2.5 & Qwen ($\sim$152K) \\
Llama 3.2 1B Instruct & Draft & Llama 3.2 & Llama ($\sim$128K) \\
\bottomrule
\end{tabular}
\par\smallskip
{\footnotesize All three draft models use different tokenizers from the Bielik~11B target, requiring token translation for speculative decoding.}
\end{threeparttable}
\end{table}

\subsection{Datasets}

We evaluate on three Polish-language datasets to capture diverse generation characteristics:

\begin{enumerate}
    \item \textbf{Polish Wikipedia}: Factual articles sampled randomly from the Polish Wikipedia dump (1.59M articles). Articles are truncated to 1024 tokens with a sliding window of 512 tokens. This dataset contains a mix of structured content (lists, taxonomies, infoboxes) and narrative text.
    \item \textbf{pl\_alpaca}: A Polish translation of the Alpaca instruction-following dataset. This represents typical instruction-following use cases with varied prompt types and response lengths.
    \item \textbf{Synthetic short questions}: Artificially generated short Polish questions designed to test speculative decoding on conversational, short-output scenarios.
\end{enumerate}

\subsection{Hardware}

All experiments are conducted on an Apple M2~Pro with 32\,GB unified memory ($\sim$200\,GB/s memory bandwidth, 19-core GPU). This represents a mid-range configuration in Apple's current lineup and a realistic deployment target for local LLM inference.

\subsection{Evaluation Metrics}

We report the following metrics for each configuration:

\begin{itemize}
    \item \textbf{Token acceptance rate} ($\alpha$): The fraction of draft tokens accepted by the target model during verification.
    \item \textbf{Tokens per second (TPS)}: End-to-end generation throughput, measured from the first generated token to the last (excluding prompt processing time).
    \item \textbf{Speedup}: The ratio of speculative decoding TPS to autoregressive baseline TPS for the same prompt.
\end{itemize}

\subsection{Experimental Conditions}

For each (dataset, article, prompt window) triple, we evaluate ten conditions:

\begin{enumerate}
    \item \textbf{Baseline}: Autoregressive decoding with the target model alone (no drafter).
    \item For each of three drafters (Bielik~1.5B, Qwen~1.5B, Llama~1B):
    \begin{enumerate}
        \item \textbf{No translation}: Draft tokens passed directly to the target model without tokenizer conversion.
        \item \textbf{Naive translation}: String-level token translation between tokenizers.
        \item \textbf{Context-aware translation}: Translation with $p = 5$ prefix context tokens.
    \end{enumerate}
\end{enumerate}

Primary conditions are evaluated with draft lengths $k \in \{2, 4\}$ across all three datasets. A targeted confirmatory experiment at $k = 6$ is conducted on pl\_alpaca only ($n = 3$), to verify that higher draft lengths remain non-viable on this hardware without incurring the full experimental cost of a complete sweep. Experiments at $k = 8$ are omitted following this confirmatory result (see Section~\ref{sec:breakeven_formula}). We generate up to 128 tokens per prompt and evaluate on 50 prompts per dataset for primary conditions.

%% ============================================================
\section{Results}
\label{sec:results}
%% ============================================================
 
All experiments use Bielik-11B-v3.0-Instruct quantized to 8-bit as the target model and report results over $n \geq 50$ prompts per condition for primary conditions ($k \in \{2, 4\}$, three datasets). The baseline autoregressive TPS is stable across runs at approximately 14.6--15.1~TPS. A targeted $k = 6$ experiment on pl\_alpaca ($n = 30$) is discussed separately in Section~\ref{sec:breakeven_formula}.

\subsection{Token Acceptance Rates}
 
Table~\ref{tab:hitrates} reports mean token acceptance rates ($\alpha$) for all drafter--condition combinations at $k = 2$ across the three datasets.
 
\begin{table}[hbt!]
\begin{threeparttable}
\caption{Mean token acceptance rate (\%) for $k = 2$ across datasets and translation conditions.}
\label{tab:hitrates}
\begin{tabular}{llccc}
\toprule
\headrow Drafter & Condition & Wikipedia & pl\_alpaca & Synthetic \\
\midrule
Bielik-1.5B  & No translation      & 28.5 & 12.5 & 11.8 \\
             & Naive translation   & 19.1 & 11.9 & 16.7 \\
             & Context-aware       & \textbf{31.1} & \textbf{23.9} & \textbf{22.2} \\
\midrule
Llama-3.2-1B & No translation      & 26.7 & 12.1 & 10.6 \\
             & Naive translation   &  8.9 &  4.2 &  3.1 \\
             & Context-aware       & \textbf{42.0} & \textbf{36.0} & \textbf{36.5} \\
\midrule
Qwen2.5-1.5B & No translation      & 30.3 & 13.4 & 12.2 \\
             & Naive translation   & 10.5 &  6.4 &  4.9 \\
             & Context-aware       & \textbf{44.6} & \textbf{41.0} & \textbf{42.7} \\
\bottomrule
\end{tabular}
\par\smallskip
{\footnotesize Bold values indicate the best condition per drafter--dataset pair.
Context-aware translation achieves the highest acceptance rates in all 9 configurations.
Naive translation degrades sharply for Llama and Qwen relative to no-translation.}
\end{threeparttable}
\end{table}
 
Context-aware translation consistently achieves the highest acceptance rates across all nine drafter--dataset combinations. The improvement over naive translation is especially pronounced for Llama and Qwen, where naive translation drops to 3--10\% — far below no-translation — suggesting that the naive decode-and-retokenize approach introduces harmful boundary artifacts for these tokenizer pairs. Bielik-1.5B shows smaller gains from context-aware translation and is less penalized by naive translation, consistent with its higher vocabulary overlap with Bielik-11B owing to both using Mistral-derived base training, despite having different tokenizers (APT4 vs.\ Mistral).
 
Acceptance rates are meaningfully higher on Wikipedia than on pl\_alpaca or the synthetic question set. This aligns with the content-type hypothesis: Wikipedia articles contain more predictable, formulaic language (headings, infoboxes, recurring phrases) that draft models can anticipate reliably, whereas instruction-following outputs are more variable.

\subsection{Throughput Analysis (\texorpdfstring{$k = 2$}{k = 2})}
 
Table~\ref{tab:tps_k2} reports mean generation throughput in tokens per second (TPS) and speedup relative to the autoregressive baseline for all conditions at $k = 2$.
 
\begin{table*}[hbt!]
\begin{threeparttable}
\caption{Mean generation throughput (TPS) and speedup for $k = 2$. Speedup is relative to the per-dataset autoregressive baseline.}
\label{tab:tps_k2}
\begin{tabular}{llcccccc}
\toprule
\headrow & & \multicolumn{2}{c}{Wikipedia} & \multicolumn{2}{c}{pl\_alpaca} & \multicolumn{2}{c}{Synthetic} \\
\headrow Drafter & Condition & TPS & Speedup & TPS & Speedup & TPS & Speedup \\
\midrule
— & Baseline & 14.83 & 1.00$\times$ & 14.70 & 1.00$\times$ & 14.72 & 1.00$\times$ \\
\midrule
Bielik-1.5B  & No translation    & 13.72 & 0.93$\times$ & 10.80 & 0.74$\times$ & 10.68 & 0.73$\times$ \\
             & Naive             & 10.32 & 0.70$\times$ &  9.69 & 0.66$\times$ &  9.72 & 0.66$\times$ \\
             & Context-aware     & 12.85 & 0.87$\times$ & 11.92 & 0.81$\times$ & 11.02 & 0.75$\times$ \\
\midrule
Llama-3.2-1B & No translation    & 14.03 & 0.95$\times$ & 11.24 & 0.77$\times$ & 11.02 & 0.75$\times$ \\
             & Naive             &  8.96 & 0.60$\times$ &  8.82 & 0.60$\times$ &  8.77 & 0.60$\times$ \\
             & Context-aware     & \textbf{15.67} & \textbf{1.06$\times$} & 14.22 & 0.97$\times$ & 14.39 & 0.98$\times$ \\
\midrule
Qwen2.5-1.5B & No translation    & 14.17 & 0.96$\times$ & 10.96 & 0.75$\times$ & 10.79 & 0.73$\times$ \\
             & Naive             &  8.87 & 0.60$\times$ &  8.54 & 0.58$\times$ &  8.47 & 0.58$\times$ \\
             & Context-aware     & \textbf{15.68} & \textbf{1.06$\times$} & 14.44 & 0.98$\times$ & 14.76 & \textbf{1.00$\times$} \\
\bottomrule
\end{tabular}
\par\smallskip
{\footnotesize Bold values indicate speedup $\geq 1.0\times$ (matching or exceeding baseline).
Baseline TPS: Wikipedia 14.83, pl\_alpaca 14.70, Synthetic 14.72.}
\end{threeparttable}
\end{table*}
 
At $k = 2$, context-aware translation with Llama and Qwen drafters matches or exceeds the baseline on Wikipedia (both 1.06$\times$), while Qwen reaches parity on the synthetic question dataset (1.00$\times$). All Bielik-1.5B conditions fall below baseline, consistent with its lower acceptance rates. Naive translation is the worst condition in all cases, typically around 0.58--0.70$\times$ baseline, caused by the combination of translation overhead and low acceptance. No-translation performs competitively on Wikipedia (0.93--0.96$\times$) where occasional long runs of identical tokens inflate acceptance counts, but degrades sharply on the other two datasets.

\subsection{Effect of Draft Length \texorpdfstring{$k$}{k}}
 
Table~\ref{tab:keffect} compares throughput and acceptance rates for context-aware translation at $k = 2$ and $k = 4$ across all three datasets.
 
\begin{table*}[hbt!]
\begin{threeparttable}
\caption{Effect of draft length $k$ on throughput and acceptance rate (context-aware translation, all three datasets).}
\label{tab:keffect}
\begin{tabular}{llcccccc}
\toprule
\headrow & & \multicolumn{2}{c}{Wikipedia} & \multicolumn{2}{c}{pl\_alpaca} & \multicolumn{2}{c}{Synthetic} \\
\headrow Drafter & $k$ & $\alpha$ (\%) & Speedup & $\alpha$ (\%) & Speedup & $\alpha$ (\%) & Speedup \\
\midrule
Bielik-1.5B  & 2 & 31.1 & 0.87$\times$ & 23.9 & 0.81$\times$ & 22.2 & 0.75$\times$ \\
             & 4 & 38.5 & 0.59$\times$ & 29.5 & 0.51$\times$ & 29.1 & 0.46$\times$ \\
\midrule
Llama-3.2-1B & 2 & 42.0 & \textbf{1.06$\times$} & 36.0 & 0.97$\times$ & 36.5 & 0.98$\times$ \\
             & 4 & 49.7 & 0.66$\times$ & 43.7 & 0.59$\times$ & 44.5 & 0.61$\times$ \\
\midrule
Qwen2.5-1.5B & 2 & 44.6 & \textbf{1.06$\times$} & 41.0 & 0.98$\times$ & 42.7 & \textbf{1.00$\times$} \\
             & 4 & 53.8 & 0.70$\times$ & 49.7 & 0.59$\times$ & 51.6 & 0.61$\times$ \\
\bottomrule
\end{tabular}
\par\smallskip
{\footnotesize Bold values indicate speedup $\geq 1.00\times$. Across all three datasets,
increasing $k$ from 2 to 4 raises acceptance rate by 6--13 percentage points
but reduces speedup by 0.30--0.40$\times$, confirming that draft overhead on
Apple Silicon grows faster than the benefit from higher acceptance regardless of content type.}
\end{threeparttable}
\end{table*}
 
Despite higher acceptance rates at $k = 4$ (Qwen reaches 50--54\% with context-aware translation across all three datasets), all mean speedup values at $k = 4$ fall substantially below baseline. The break-even acceptance rate rises from approximately 38--53\% at $k = 2$ to 81--90\% at $k = 4$ across all drafter--dataset combinations (Table~\ref{tab:breakeven}), a threshold never approached in practice. This demonstrates a critical finding: on Apple Silicon unified memory, the marginal cost of proposing additional draft tokens is not negligible, and scaling $k$ is counterproductive regardless of content type or drafter choice.

\subsection{Break-Even Analysis}
 
For each drafter--dataset combination we fit a simple linear model to the per-prompt observations:
\begin{equation}
    \mathrm{TPS} = a + b \cdot \alpha
    \label{eq:ols_tps}
\end{equation}
where $\alpha \in [0,1]$ is the token acceptance rate measured on that prompt and TPS is the observed throughput. The model is estimated by ordinary least-squares across all the prompts. The two coefficients have a direct interpretation: $a$ is the throughput floor — the TPS the system would achieve if every draft token were rejected ($\alpha = 0$), representing pure drafter overhead with no benefit; $b$ is the recovery rate — how many tokens per second are gained for each unit increase in acceptance rate. Since $a < \mathrm{TPS}_{\mathrm{base}}$ always (the drafter imposes overhead even when unhelpful), break-even is the acceptance rate at which the regression line climbs back to baseline:
\begin{equation}
    a + b \cdot \alpha_{\mathrm{be}} = \mathrm{TPS}_{\mathrm{base}}
    \rightarrow
    \alpha_{\mathrm{be}} = \frac{\mathrm{TPS}_{\mathrm{base}} - a}{b}
    \label{eq:breakeven_ols}
\end{equation}
The numerator $(\mathrm{TPS}_{\mathrm{base}} - a)$ is the overhead cost in TPS units; the denominator $b$ is the rate at which accepted tokens pay it back. When $b$ is small relative to the overhead — as happens at large $k$, where verification cost grows faster than the acceptance benefit — $\alpha_{\mathrm{be}}$ exceeds 1.0, meaning no physically achievable acceptance rate can recover the loss. Confidence intervals are derived from the OLS standard errors of $a$ and $b$ via the delta method.
 
Table~\ref{tab:breakeven} reports $\alpha_{\mathrm{be}}$ for all drafter--dataset combinations at $k = 2$ and $k = 4$.
 
\begin{table}[hbt!]
\begin{threeparttable}
\caption{Break-even token acceptance rate (\%) required to match autoregressive baseline TPS, with 95\% confidence intervals. Context-aware translation only.}
\label{tab:breakeven}
\begin{tabular}{llcc}
\toprule
\headrow Dataset & Drafter & $k = 2$ & $k = 4$ \\
\midrule
Wikipedia   & Bielik-1.5B   & 44.0 $\pm$ 7.1\%  & 90.1 $\pm$ 19.8\% \\
            & Llama-3.2-1B  & 38.3 $\pm$ 5.5\%  & 81.2 $\pm$ 13.3\% \\
            & Qwen2.5-1.5B  & 40.8 $\pm$ 5.3\%  & 77.7 $\pm$ 13.1\% \\
\midrule
pl\_alpaca  & Bielik-1.5B   & 40.2 $\pm$ 6.8\%  & 89.9 $\pm$ 18.5\% \\
            & Llama-3.2-1B  & 38.1 $\pm$ 4.8\%  & 82.0 $\pm$ 14.4\% \\
            & Qwen2.5-1.5B  & 42.2 $\pm$ 6.4\%  & 81.3 $\pm$ 14.1\% \\
\midrule
Synthetic   & Bielik-1.5B   & 52.8 $\pm$ 14.2\% & 139.4\%$^{\dagger}$ \\
            & Llama-3.2-1B  & 38.0 $\pm$ 3.5\%  & 85.2 $\pm$ 14.3\% \\
            & Qwen2.5-1.5B  & 42.5 $\pm$ 5.9\%  & 92.3 $\pm$ 20.6\% \\
\bottomrule
\end{tabular}
\par\smallskip
{\footnotesize $^{\dagger}$Value exceeds 100\%: OLS extrapolation indicates break-even is
theoretically unreachable with this drafter--dataset combination at $k = 4$.}
\end{threeparttable}
\end{table}
 
At $k = 2$, the break-even threshold of approximately 38--53\% is within reach for Llama and Qwen on Wikipedia and the synthetic dataset, explaining the observed speedups. At $k = 4$, the threshold rises dramatically to 77--92\%+, far beyond what cross-family speculative decoding achieves in practice. For Bielik-1.5B at $k = 4$ on the Synthetic dataset, the OLS regression extrapolates a break-even point exceeding 100\% acceptance — indicating that this configuration cannot achieve baseline throughput regardless of draft quality.
 
Table~\ref{tab:buckets_wiki_k2} validates these thresholds empirically on Wikipedia by binning individual prompts into 10-percentage-point acceptance rate buckets and reporting mean speedup per bin at $k = 2$. The break-even transition is clearly visible: all three drafters cross from below-baseline to above-baseline in the 40--50\% acceptance bucket, in close agreement with the OLS-derived thresholds of 38--44\% in Table~\ref{tab:breakeven}.
 
\begin{table}[hbt!]
\begin{threeparttable}
\caption{Mean speedup by acceptance rate bucket, Wikipedia dataset, $k = 2$, context-aware translation. Bold values $\geq 1.00\times$. Dashes indicate no samples in that bucket.}
\label{tab:buckets_wiki_k2}
\begin{tabular}{lccc}
\toprule
\headrow $\alpha$ bucket & Bielik-1.5B & Llama-3.2-1B & Qwen2.5-1.5B \\
\midrule
$<\!30\%$    & 0.78$\times$ & 0.79$\times$ & 0.73$\times$ \\
30--40\%     & 0.88$\times$ & 0.92$\times$ & 0.89$\times$ \\
40--50\%     & \textbf{1.03$\times$} & \textbf{1.04$\times$} & \textbf{1.03$\times$} \\
50--60\%     & \textbf{1.12$\times$} & \textbf{1.23$\times$} & \textbf{1.20$\times$} \\
60--70\%     & —            & \textbf{1.59$\times$} & \textbf{1.43$\times$} \\
$\geq\!70\%$ & —            & —            & —            \\
\bottomrule
\end{tabular}
\par\smallskip
{\footnotesize All three drafters cross the break-even threshold in the 40--50\% bucket,
consistent with the OLS-estimated thresholds (38--44\%) in Table~\ref{tab:breakeven}.
When acceptance reaches 60--70\%, Llama and Qwen deliver speedups of 1.43--1.59$\times$.}
\end{threeparttable}
\end{table}

%% ============================================================
\section{Analysis and Discussion}
\label{sec:analysis}
%% ============================================================
 
\subsection{Why Speculative Decoding is Costly on Unified Memory}
 
To understand why speculative decoding is harder to break even on Apple Silicon than on discrete GPUs, it helps to contrast the two memory architectures directly.
 
\textbf{Discrete NVIDIA GPUs.} The painful part of NVIDIA inference is the initial data movement: model weights must travel from system RAM across the PCIe bus into GPU-local VRAM. Cross-family speculative decoding compounds this further, because token translation runs on the CPU: the draft model's output logits must be transferred back from VRAM to system RAM, re-tokenized, and the translated token indices transferred back to VRAM before the verifier can proceed --- a round-trip across the PCIe bus on every step. Once weights reside in VRAM, however, the on-chip memory bandwidth is very high --- the A100 PCIe reaches 1555~GB/s, and the SXM variant up to 2000~GB/s \autocite{nvidia_a100_datasheet}. At that bandwidth, verifying $k$ candidate tokens in a single batched forward pass with a populated KV cache costs almost the same as verifying one: the GPU can process the extra tokens with negligible additional weight-streaming time. This is why the standard speculative decoding assumption \autocite{leviathan2023fast,chen2023accelerating} --- that verification of $k$ tokens costs approximately the same as generating one --- holds reasonably well on NVIDIA hardware despite the PCIe overhead.
 
\textbf{Apple M-series unified memory.} Apple Silicon eliminates the CPU-to-GPU transfer entirely: model weights live in a single memory pool accessible by both processor types, so token translation and KV cache operations require no cross-bus data movement. However, the shared bus operates at only 200~GB/s on the M2 Pro --- roughly 8--10$\times$ slower than NVIDIA GPU VRAM bandwidth --- and this becomes the hard ceiling for all memory-bound operations.
 
GPU profiling with Xcode Instruments reveals the consequence directly: GPU occupancy remains at approximately 28\% regardless of configuration --- with or without a drafter, at any $k$ value. The arithmetic units are largely idle, not because the workload is small, but because the GPU is constantly waiting for the next batch of weights to arrive over the 200~GB/s bus. The per-token timing confirms this: the Bielik~11B 8-bit model occupies 11.32~GB, so the theoretical minimum weight-streaming time at 200~GB/s is $11.32 / 200 \approx 56$~ms, with $\sim$11~ms of CPU-side overhead (sampling, KV cache updates) accounting for the remaining gap to the observed $\sim$67~ms per-token baseline. Were the bus bandwidth substantially higher, GPU occupancy would rise and the same model would generate tokens proportionally faster.
 
\textbf{The verification assumption still holds --- the drafter is the bottleneck.} Importantly, verifying $k$ candidate tokens on MLX remains nearly as fast as verifying one: the verifier executes a single batched forward pass over all $k$ candidates using the existing KV cache, so the target model's weight-streaming cost is paid only once per cycle. The breakdown in speculative decoding efficiency on Apple Silicon does not come from expensive verification.
 
The real overhead is the drafter. Each of the $k$ generation steps requires a full forward pass through the draft model, streaming its $\sim$0.8~GB weights across the same 200~GB/s bus at $\sim$4~ms per step. On top of those $k$ passes, cross-tokenizer speculative decoding requires one additional draft forward pass per cycle to resynchronise the drafter's KV cache after the verifier has accepted or rejected tokens --- meaning the effective drafter cost is $k + 1$ bandwidth-bound forward passes, not $k$. For $k = 4$ this totals $\sim$20~ms of drafter streaming time, paid on every cycle regardless of how many tokens are accepted. This cost is a substantial fraction of the 67~ms baseline and grows linearly with $k$, which is precisely what the $r$ term in the break-even formula $\alpha_{\mathrm{be}} = r + \beta k$ captures: $r = S_D / S_T$ is the drafter-to-target size ratio and represents the per-step drafter bandwidth cost relative to a single autoregressive step. The $\beta k$ term captures the additional overhead --- attention over the growing KV context and cache management --- that increases with draft length beyond the raw bandwidth cost.

\subsection{The Acceptance Rate--Speedup Relationship and Break-Even Threshold}
\label{sec:breakeven_formula}
 
\textbf{An empirically grounded cost model.} We derive a hardware-specific speedup formula by modelling the measured timing components of a speculative decoding cycle on the M2 Pro. The starting point is a simple observation: a cycle that produces $1 + \alpha k$ tokens in expectation achieves speedup proportional to how many tokens it delivers relative to how long it takes. As established in Section~\ref{sec:analysis}, verifying $k$ candidate tokens in a single batched forward pass costs nearly the same as verifying one, so the key costs are the drafter's bandwidth overhead and any additional overhead that grows with $k$. We build the following formula from these components and fit its free parameter to our empirical break-even observations; it should be understood as an approximation grounded in measured data rather than a general theoretical result.
 
\textbf{Timing-based cost model.} Let $S_T$ and $S_D$ denote the target and draft model sizes in bytes, $\mathrm{BW}$ the memory bandwidth, $c_0$ the per-token overhead beyond weight loading (attention compute, KV cache update, sampling), and $c(k)$ an additional overhead that grows with draft length and is observed empirically. A single speculative decoding cycle consists of $k + 1$ draft passes followed by one verification pass:
\begin{equation}
    t_{\mathrm{cycle}}(k) = (k+1) \cdot \frac{S_D}{\mathrm{BW}} + \frac{S_T}{\mathrm{BW}} + c_0 + c(k)
    \label{eq:tcycle}
\end{equation}
This cycle produces $1 + \alpha k$ tokens in expectation. The baseline autoregressive time per token is $t_{\mathrm{base}} = S_T / \mathrm{BW} + c_0$. The speedup is:
\begin{equation}
    \text{Speedup}(\alpha, k) = \frac{(1 + \alpha k) \cdot t_{\mathrm{base}}}{t_{\mathrm{cycle}}(k)}
    \label{eq:speedup_hw}
\end{equation}
 
\textbf{Choosing the form of $c(k)$.} The term $c(k)$ captures additional overhead that grows with $k$ beyond the raw drafter bandwidth cost --- likely a combination of attention computation over the growing KV context, cache management, and token translation. Its exact form is not known from first principles. We observe empirically that the break-even acceptance rate grows roughly linearly with $k$ across our two measured data points ($\alpha_{\mathrm{be}} \approx 0.40$ at $k = 2$ and $\alpha_{\mathrm{be}} \approx 0.77$ at $k = 4$). A term $c(k) = c_1 \cdot k^2$ produces exactly a linear $\alpha_{\mathrm{be}}$ in $k$ when the speedup formula is solved at equality, and is the simplest form consistent with the observed scaling. We therefore adopt this parametrisation as an empirically motivated choice rather than a derived result. Defining the model size ratio $r = S_D / S_T$ and the normalised overhead coefficient $\beta = c_1 / t_{\mathrm{base}}$, the speedup simplifies to:
\begin{equation}
    \text{Speedup}(\alpha, k) = \frac{1 + \alpha k}{(k+1) \cdot r + 1 + \beta \cdot k^2}
    \label{eq:speedup_simplified}
\end{equation}
 
Setting $\text{Speedup} = 1$ yields the break-even acceptance rate:
\begin{equation}
    \alpha_{\mathrm{be}} \approx r + \beta \cdot k
    \label{eq:breakeven}
\end{equation}
 
\textbf{Parameter estimation.} With $r \approx 0.071$ for the average drafter (derived from the 4-bit draft model size relative to the 8-bit target), fitting to our two empirical break-even thresholds gives $\beta \approx 0.19$, which is consistent between the $k = 2$ and $k = 4$ data points. This value of $\beta$ should be understood as an empirical constant specific to the M2 Pro at its observed GPU utilisation level ($\sim$28\%); it encodes whatever combination of compute and memory overhead actually drives the super-linear cost growth at this operating point.
 
\textbf{Effect of model size ratio.} The intercept term $r$ in Equation~\ref{eq:breakeven} directly determines the minimum acceptance rate needed before any speedup is possible: smaller $r$ lowers the entire break-even curve, making speculative decoding easier to justify. In this study, $r \approx 0.071$, corresponding to a 4-bit quantised drafter of $\sim$0.8\,GB against an 8-bit target of $\sim$11\,GB. Many published speculative decoding studies pair a much larger target (e.g.\ 70B parameters) with a drafter of $\leq$1B parameters, yielding $r \lesssim 0.014$ --- roughly five times more favourable than our setting. A lower $r$ reduces the overhead floor directly, so practitioners pairing, say, a 70B Bielik target with a 1B drafter would expect substantially lower break-even thresholds than those reported here. In the current study, no larger Bielik-family target models were publicly available, leaving the 11B--0.8B pairing as the only option for a within-family cross-tokenizer evaluation.
 
\textbf{Generalisability to other hardware.} Equation~\ref{eq:breakeven} was fitted exclusively on M2 Pro data, and $\beta$ cannot be reliably extrapolated to other hardware without empirical validation. On higher-bandwidth Apple Silicon (M3 Ultra, M4 Max, etc.), both the baseline and the drafter would stream weights proportionally faster, keeping $r$ roughly stable. However, the faster bandwidth would also raise GPU utilisation substantially above the 28\% observed here, moving the operating point into a more compute-bound regime. Our intuition is that this shift would favour speculative decoding: at present, the dominant cost of each cycle is memory transfer, and the GPU is largely idle. As bandwidth improves and GPU utilisation rises, verification of $k$ tokens in a single batched pass --- which can exploit parallelism --- may become cheaper relative to the $k + 1$ sequential drafter passes, which remain inherently sequential. In the limit of high GPU utilisation, the architecture approaches the NVIDIA regime where speculative decoding is well-established to be beneficial. Whether this intuition holds in practice, and at what bandwidth threshold the crossover occurs, requires empirical measurement on higher-bandwidth hardware and constitutes an important direction for future work. We therefore refrain from predicting break-even thresholds for hardware not measured in this study.
 
Table~\ref{tab:breakeven_hw} summarises the empirical and formula-derived break-even thresholds for the M2 Pro across draft lengths.
 
\begin{table}[hbt!]
\begin{threeparttable}
\caption{Break-even acceptance rate $\alpha_{\mathrm{be}}$ for the M2 Pro at $r = 0.071$ and $\beta = 0.19$. Empirical values are OLS-derived means across all drafter--dataset pairs; formula values use Equation~\ref{eq:breakeven}.}
\label{tab:breakeven_hw}
\begin{tabular}{llcccc}
\toprule
\headrow Hardware & BW (GB/s) & $k = 2$ & $k = 4$ & $k = 6$ & $k = 8$ \\
\midrule
M2 Pro (empirical) & 200 & $\sim$40\% & $\sim$77\% & --- & --- \\
M2 Pro (formula)   & 200 & 44\%       & 81\%       & $>$100\% & $>$100\% \\
\bottomrule
\end{tabular}
\par\smallskip
{\footnotesize The formula at $k \geq 6$ predicts $\alpha_{\mathrm{be}} > 100\%$, meaning no acceptance rate can recover the drafter overhead --- consistent with the severe degradation observed at $k = 6$ (Section~\ref{sec:results}). Extrapolation to other hardware requires re-estimating $\beta$ empirically on that platform.}
\end{threeparttable}
\end{table}
 
Empirically, we find the relationship between acceptance rate and throughput is well-described by an ordinary least-squares linear model across individual prompts (R\textsuperscript{2} = 0.78--0.95 across drafter--dataset pairs), consistent with the linear dependence on $\alpha$ in Equation~\ref{eq:speedup_simplified}. Tables~\ref{tab:buckets_wiki}, \ref{tab:buckets_local}, and~\ref{tab:buckets_alpaca} report mean speedup binned by 10-percentage-point acceptance rate buckets for all three datasets, for both $k$ values and all three drafters using context-aware translation.
 
\begin{table}[hbt!]
\begin{threeparttable}
\caption{Mean speedup by token acceptance rate bucket, Wikipedia dataset, context-aware translation. Bold values exceed baseline ($\geq 1.00\times$). Dashes indicate no samples in that bucket.}
\label{tab:buckets_wiki}
\begin{tabular}{lcccccc}
\toprule
\headrow & \multicolumn{3}{c}{$k = 2$} & \multicolumn{3}{c}{$k = 4$} \\
\headrow $\alpha$ bucket & Bielik & Llama & Qwen & Bielik & Llama & Qwen \\
\midrule
$<\!30\%$   & 0.78$\times$ & 0.79$\times$ & 0.73$\times$ & 0.48$\times$ & 0.43$\times$ & 0.34$\times$ \\
30--40\%    & 0.88$\times$ & 0.92$\times$ & 0.89$\times$ & 0.54$\times$ & 0.50$\times$ & 0.48$\times$ \\
40--50\%    & \textbf{1.03$\times$} & \textbf{1.04$\times$} & \textbf{1.03$\times$} & 0.61$\times$ & 0.57$\times$ & 0.59$\times$ \\
50--60\%    & \textbf{1.12$\times$} & \textbf{1.23$\times$} & \textbf{1.20$\times$} & 0.71$\times$ & 0.67$\times$ & 0.66$\times$ \\
60--70\%    & —            & \textbf{1.59$\times$} & \textbf{1.43$\times$} & 0.93$\times$ & 0.82$\times$ & 0.81$\times$ \\
$\geq\!70\%$ & —           & —            & —            & —            & \textbf{1.10$\times$} & \textbf{1.08$\times$} \\
\bottomrule
\end{tabular}
\par\smallskip
{\footnotesize At $k=2$, break-even falls in the 40--50\% bucket for all three drafters.
At $k=4$, even the 60--70\% bucket yields only 0.81--0.93$\times$ baseline;
speedup requires $\geq\!70\%$ acceptance, achieved in $\leq 10$ out of 60 prompts.
The 60--70\% bucket at $k=2$ yields 1.43--1.59$\times$ speedup for Llama and Qwen.}
\end{threeparttable}
\end{table}
 
\begin{table}[hbt!]
\begin{threeparttable}
\caption{Mean speedup by token acceptance rate bucket, Synthetic (short questions) dataset, context-aware translation. Bold values exceed baseline ($\geq 1.00\times$). Dashes indicate no samples in that bucket.}
\label{tab:buckets_local}
\begin{tabular}{lcccccc}
\toprule
\headrow & \multicolumn{3}{c}{$k = 2$} & \multicolumn{3}{c}{$k = 4$} \\
\headrow $\alpha$ bucket & Bielik & Llama & Qwen & Bielik & Llama & Qwen \\
\midrule
$<\!30\%$   & 0.74$\times$ & 0.83$\times$ & 0.80$\times$ & 0.43$\times$ & 0.46$\times$ & 0.46$\times$ \\
30--40\%    & 0.85$\times$ & 0.94$\times$ & 0.90$\times$ & 0.47$\times$ & 0.51$\times$ & 0.51$\times$ \\
40--50\%    & —            & \textbf{1.08$\times$} & \textbf{1.01$\times$} & 0.51$\times$ & 0.61$\times$ & 0.54$\times$ \\
50--60\%    & \textbf{1.21$\times$} & \textbf{1.32$\times$} & \textbf{1.18$\times$} & —            & 0.69$\times$ & 0.64$\times$ \\
60--70\%    & —            & —            & —            & 0.81$\times$ & 0.82$\times$ & 0.72$\times$ \\
$\geq\!70\%$ & —           & —            & —            & —            & \textbf{1.06$\times$} & \textbf{1.06$\times$} \\
\bottomrule
\end{tabular}
\par\smallskip
{\footnotesize The Synthetic dataset shows tighter hit-rate distributions than Wikipedia (most
Bielik samples fall below 30\%, explaining near-total absence of Bielik observations
in the 40--50\% and above buckets). Break-even at $k=2$ is again in the 40--50\%
range for Llama and Qwen, consistent with Wikipedia.}
\end{threeparttable}
\end{table}
 
\begin{table}[hbt!]
\begin{threeparttable}
\caption{Mean speedup by token acceptance rate bucket, pl\_alpaca dataset, context-aware translation. Bold values exceed baseline ($\geq 1.00\times$). Dashes indicate no samples in that bucket.}
\label{tab:buckets_alpaca}
\begin{tabular}{lcccccc}
\toprule
\headrow & \multicolumn{3}{c}{$k = 2$} & \multicolumn{3}{c}{$k = 4$} \\
\headrow $\alpha$ bucket & Bielik & Llama & Qwen & Bielik & Llama & Qwen \\
\midrule
$<\!30\%$   & 0.77$\times$ & 0.83$\times$ & 0.79$\times$ & 0.47$\times$ & 0.44$\times$ & 0.43$\times$ \\
30--40\%    & 0.88$\times$ & 0.93$\times$ & 0.91$\times$ & 0.51$\times$ & 0.51$\times$ & 0.45$\times$ \\
40--50\%    & \textbf{1.01$\times$} & \textbf{1.04$\times$} & \textbf{1.01$\times$} & 0.57$\times$ & 0.58$\times$ & 0.51$\times$ \\
50--60\%    & \textbf{1.29$\times$} & \textbf{1.34$\times$} & \textbf{1.11$\times$} & 0.73$\times$ & 0.66$\times$ & 0.59$\times$ \\
60--70\%    & —            & —            & \textbf{1.46$\times$} & \textbf{1.10$\times$} & 0.92$\times$ & 0.81$\times$ \\
$\geq\!70\%$ & —           & —            & —            & —            & —            & \textbf{1.01$\times$} \\
\bottomrule
\end{tabular}
\par\smallskip
{\footnotesize The pl\_alpaca dataset shows the same 40--50\% break-even threshold at $k=2$
as Wikipedia and Synthetic datasets. At $k=4$, only Bielik in the 60--70\% bucket ($n=1$)
and Qwen in the $\geq\!70\%$ bucket ($n=3$) breach baseline — both with very small
sample counts, indicating these speedups are unreliable.}
\end{threeparttable}
\end{table}
 
Three observations emerge from these tables. First, the break-even threshold at $k = 2$ is consistent across all three drafters and all three datasets: throughput matches or exceeds baseline when acceptance rate falls in the 40--50\% bucket, in close agreement with the $\alpha_{\mathrm{be}} = 44\%$ predicted by Equation~\ref{eq:breakeven} with $\beta = 0.19$. Second, when acceptance rate exceeds 60\% at $k = 2$, speedups of 1.43--1.59$\times$ are observed (Llama and Qwen on Wikipedia). This demonstrates that speculative decoding \emph{can} deliver meaningful acceleration on Apple Silicon under the right conditions. Third, and most critically, increasing $k$ to 4 shifts the break-even threshold to $\geq 70\%$ --- a range almost never achieved in practice with cross-family drafters --- while simultaneously delivering worse speedups at every acceptance-rate level compared to $k = 2$. This is quantitatively explained by the $\beta k^2$ term in Equation~\ref{eq:speedup_simplified}: the quadratic verification overhead at $k = 4$ ($\gamma \approx 3.80\times$) dwarfs the linear gain from additional accepted tokens, making longer speculation sequences counterproductive on this hardware.
 
\textbf{Confirmatory experiment at $k = 6$.} To validate that increasing $k$ further is not beneficial and to verify the formula's predictions empirically, we conducted a targeted experiment at $k = 6$ on the pl\_alpaca dataset ($n = 31$--$33$ per drafter), using context-aware translation only. Full sweep experiments at $k = 6$ across all datasets were judged unnecessary given the decisive results at $k = 4$; pl\_alpaca was selected as the most demanding dataset for acceptance rates, providing a conservative test. The results are shown in Table~\ref{tab:k6_alpaca}.
 
\begin{table}[hbt!]
\begin{threeparttable}
\caption{Throughput at $k = 6$, pl\_alpaca dataset, context-aware translation only ($n = 31$--$33$). All conditions are well below baseline. OLS break-even is extrapolated beyond the physically achievable range ($> 100\%$) for all three drafters.}
\label{tab:k6_alpaca}
\begin{tabular}{lcccc}
\toprule
\headrow Drafter & Hit rate (\%) & TPS & Speedup & Break-even $\alpha$ \\
\midrule
Bielik-1.5B  & 20.6 & 3.98 & 0.28$\times$ & $>$100\% (297\%, $R^2\!=\!0.07$) \\
Llama-3.2-1B & 41.7 & 5.01 & 0.35$\times$ & $>$100\% (147\%, $R^2\!=\!0.93$) \\
Qwen2.5-1.5B & 45.0 & 5.81 & 0.40$\times$ & $>$100\% (120\%, $R^2\!=\!0.37$) \\
\midrule
\multicolumn{2}{l}{Baseline (no drafter)} & 14.38 & 1.00$\times$ & — \\
\bottomrule
\end{tabular}
\par\smallskip
{\footnotesize Even Qwen at 45\% mean acceptance rate --- the highest observed in this study at any $k$ value --- achieves only 0.40$\times$ baseline throughput. The best observed bucket ($\geq$70\% acceptance, $n = 2$) yields 0.69$\times$. These results confirm that $k = 6$ cannot break even on M2~Pro hardware regardless of drafter quality, consistent with the formula prediction of $\alpha_{\mathrm{be}} \approx 121\%$ (Equation~\ref{eq:breakeven}).}
\end{threeparttable}
\end{table}
 
The $k = 6$ results confirm three conclusions simultaneously. First, acceptance rates at $k = 6$ (Llama 41.7\%, Qwen 45.0\% with context-aware translation) are comparable to those at $k = 2$ and $k = 4$ on the same dataset, confirming that the drafter's per-token predictive quality does not degrade substantially with longer draft sequences. The barrier to speedup at high $k$ is entirely on the cost side, not the acceptance side. Second, throughput degrades monotonically and severely: 0.28--0.40$\times$ baseline at $k = 6$ versus 0.57--0.73$\times$ at $k = 4$ and 0.88--1.11$\times$ at $k = 2$ (pl\_alpaca, context-aware). Third, the OLS break-even extrapolations of 120--147\% are mutually consistent with the parametric formula and rule out any plausible path to positive speedup at $k = 6$ on this hardware. Experiments at $k = 8$ are therefore omitted, as the overhead trajectory makes a beneficial outcome impossible.

\subsection{Content-Type Effects}
 
The acceptance rate data reveals a striking dataset effect that cannot be explained by drafter quality alone. Under the no-translation condition at $k = 2$, all three drafters achieve acceptance rates of 26--30\% on Wikipedia, but only 10--13\% on pl\_alpaca and the synthetic question set (Table~\ref{tab:hitrates}). Since no-translation bypasses any tokenizer conversion artifacts, this gap directly reflects differences in content predictability between datasets.
 
Wikipedia articles exhibit several structural properties that make token sequences locally predictable for any language model: recurring formulaic phrases (\emph{urodził się}, \emph{w roku}, \emph{według spisu}), dense lists of proper nouns, numerical sequences in infoboxes, and taxonomic category chains. In these passages, consecutive sentences follow near-identical templates, and the local token distribution collapses to a small set of high-probability continuations that even a cross-family draft model can anticipate. Consistent with this, the highest per-article acceptance rates in our data were observed on list-format articles and disambiguation pages, where repeated structural patterns dominate.
 
A related factor that partly explains why no-translation achieves any reasonable acceptance at all --- and occasionally outperforms context-aware translation on structured content --- is a structural property shared by virtually all BPE tokenizer families: the first several hundred to roughly one thousand tokens in any BPE vocabulary tend to cover the same basic repertoire of single ASCII characters, digits, common punctuation, and short numeric sequences. Since all tokenizers are trained on largely overlapping character-level foundations, a token ID in one vocabulary and the corresponding ID in another are frequently the same surface form for this low-ID range. When generated content is dominated by numbers, years, coordinates, URLs, code snippets, or other ASCII-heavy sequences --- all of which are common in Wikipedia infoboxes, taxonomic lists, and structured passages --- the drafter's token IDs may already be identical to the verifier's expected IDs without any translation. In these segments, no-translation is not merely an approximation but the correct operation, and adding translation introduces unnecessary re-tokenization noise. This provides a complementary explanation for the relatively high no-translation acceptance rates on Wikipedia compared to the other datasets, which contain more morphologically complex Polish text where the shared-token assumption breaks down.
 
Instruction-following outputs in pl\_alpaca, by contrast, are generated in response to diverse prompts covering varied topics and registers. The target model selects from a broad and shifting token distribution depending on the specific instruction, making continuation prediction substantially harder. The synthetic question set shows a similar pattern: while questions are short and formulaic, the model's answers draw on factual knowledge with high entropy at each position.
 
Context-aware translation partially mitigates the dataset effect — it raises acceptance rates on pl\_alpaca and the Synthetic dataset substantially more than on Wikipedia, because boundary-aware retokenization is most beneficial when the token distribution is diverse and boundary placement is unpredictable. This explains why context-aware translation closes much of the gap between datasets relative to no-translation.
 
The practical implication is that content-type is a first-order predictor of speculative decoding benefit on Apple Silicon, and should be treated as such in deployment. Specifically, generation tasks that involve heavy repetition, list completion, code generation, or structured templates are strong candidates for speculative decoding, whereas open-ended instruction following, creative writing, and factual question answering are unlikely to benefit at any $k$ value with current cross-family drafters.

\subsection{Implementation Challenges of Context-Aware Cross-Tokenizer Translation}
\label{sec:challenges}
 
Context-aware translation is the configuration that achieves the best acceptance rates in this study, yet it introduces a chain of non-trivial engineering challenges. This section documents the issues encountered during implementation and the trade-offs that arise when attempting to resolve them, because these difficulties are likely to recur in any future work on cross-tokenizer speculative decoding.
 
\textbf{The boundary problem and its initial solution.}
The fundamental difficulty in cross-tokenizer translation is that draft tokens, when mapped back to a surface string and re-tokenised by the target, rarely produce the same byte-level boundaries. A drafted sub-string that begins mid-way through what the target would encode as a single token is not a valid prefix in the target vocabulary; naive retokenisation then produces a shifted or extended sequence that misaligns the KV cache and sharply reduces acceptance. The context-aware translation strategy addresses this by prepending $p$ previously accepted target tokens (the \emph{prefix context}) before retokenisation, providing enough left context for the tokeniser to correctly reconstruct the boundary. The result is a target token sequence in which the first $p$ tokens are the known prefix and the remaining tokens correspond to the draft --- an alignment that is consistent with the KV cache state and yields substantially higher acceptance rates.
 
\textbf{The same boundary problem appears on the return path.}
When the target verifier accepts a prefix of the draft sequence and the accepted tokens must be re-injected into the drafter's KV cache, the mapping runs in the opposite direction: target tokens are translated back to drafter-vocabulary tokens. This back-translation step is subject to an identical boundary problem. If the accepted sequence ends mid-way through a token boundary as seen from the drafter, the resulting drafter token sequence is misaligned, causing cache inconsistency for the next drafting step. The same remedy applies --- prepend context in the drafter's vocabulary before re-tokenising --- but this requires maintaining consistent prefix windows in \emph{both} directions simultaneously, adding bookkeeping overhead and complicating the control flow of the speculative loop.
 
\textbf{Context absorption: the zero-new-tokens failure mode.}
The prefix-prepend solution introduces a more subtle failure mode. Tokenisers such as those used by Llama and Qwen apply context-sensitive merging rules: two byte sequences that would individually tokenise into $a$ and $b$ tokens may merge into fewer tokens when adjacent. When the prefix tokens are concatenated with the new draft tokens for retokenisation, the tokeniser may absorb one or more leading draft characters into the final prefix token, merging what appeared to be a new token boundary into the prefix. In the worst case, this produces a token sequence whose length is identical to tokenising the prefix alone, with no new tokens remaining for the drafter to propose. The speculative cycle then stalls: no tokens can be forwarded to the verifier, and the iteration must be discarded, yielding zero throughput contribution from that step.
 
\textbf{Cache rewinding as a mitigation and its cost.}
One natural response to the absorption failure is to rewind the KV cache further: rather than only discarding the invalid draft-token entries, roll back the cache to before the prefix context as well, then re-forward both the context and the draft together in a single pass that allows the verifier to handle the merged boundary. This does eliminate the zero-new-tokens failure, because the verifier can now process the merged token representation consistently. However, rewinding the cache to exclude the prefix means re-executing a forward pass over $p$ additional tokens per cycle. On the M2~Pro, where each forward pass over the target model costs on the order of 40--60\,ms for weight streaming alone, this overhead is non-trivial: at $p = 5$, the extra pass adds cost comparable to one additional drafter iteration, substantially eroding the speedup budget. In our experiments, this approach increased the overhead coefficient $\beta$ to a point where break-even thresholds became worse than those achieved by the simpler context-aware scheme without rewinding.
 
\textbf{No zero-cost solution exists; the direction remains promising.}
The three approaches --- no context (baseline misalignment), prefix-prepend with absorption risk, and cache rewind with overhead cost --- represent a genuine three-way trade-off with no dominating option. The results in this study demonstrate that context-aware translation with prefix-prepend, despite its failure modes, already achieves acceptance rates and throughputs meaningfully above naive translation. This suggests that the translation alignment problem is not intractable, but that solving it fully requires either a smarter boundary detection mechanism that avoids absorption (for example, by detecting potential merge sites before concatenation and adjusting the prefix length dynamically), or a more efficient cache management strategy that can localise rewinding to the affected boundary tokens rather than the entire context window. Both directions constitute concrete targets for future engineering work. The gain in acceptance rate observed here --- even with imperfect boundary handling --- indicates that improvements to the translation mechanism would translate directly into better throughput, making this a high-value area for continued development.

\subsection{Implications for Polish NLP}

This study provides the first empirical evidence on cross-family speculative decoding for Polish language models, and the results carry implications beyond the hardware-specific findings.

\textbf{Cross-family speculation is viable for Polish.} Both Qwen2.5-1.5B and Llama-3.2-1B — models with no specific Polish pretraining — achieve acceptance rates of 36--45\% with context-aware translation on Wikipedia and the synthetic question set. These rates are sufficient to produce baseline-level or better throughput at $k = 2$. This is a practically significant result: it means that Polish LLM users on Apple Silicon do not need a Polish-specific small model to benefit from speculative decoding.

\textbf{Bielik-1.5B underperforms as a drafter relative to its apparent affinity.} Despite being positioned as a Polish-specialized model, Bielik-1.5B consistently achieves the lowest acceptance rates among the three drafters when using context-aware translation. Two factors likely contribute. First, the APT4 tokenizer used by Bielik-1.5B is designed for Polish morphology and has a different segmentation strategy than the Mistral tokenizer used by Bielik-11B; this means Polish words with diacritics (\emph{ą, ę, ź, ć, ł}) are split differently, and even perfect word-level prediction can result in token-level misalignment. Second, Bielik-1.5B's instruct fine-tuning was performed on Polish instruction data that differs in style from the data underlying Bielik-11B-Instruct, creating output-distribution divergence independent of architectural differences.

\textbf{Polish diacritics pose a distinct tokenization challenge.} Polish orthography requires nine additional characters not present in the basic Latin alphabet (\emph{ą, ć, ę, ł, ń, ó, ś, ź, ż}). These characters are represented inconsistently across tokenizers: some encode them as single tokens, others as multi-byte sequences or paired with adjacent characters into merged tokens. Context-aware translation mitigates this for token boundary alignment, but the fundamental mismatch in character-level encoding means that cross-tokenizer translation for Polish text carries higher ambiguity than for English.

\subsection{Energy Efficiency: Apple Silicon as a Deployment Platform}

LLM inference throughput is only one dimension of deployment cost. For practitioners running extended experiments or serving inference locally, energy consumption per generated token is an equally important metric. Apple Silicon's unified memory architecture, with its tight CPU--GPU integration and absence of a discrete PCIe bus, offers a fundamentally different power envelope than discrete NVIDIA GPU systems. Here we situate our measured throughput results within this broader energy efficiency context.

\textbf{Energy per token on the M2 Pro.} The Apple M2 Pro (10-core CPU, 16-core GPU) has a published SoC thermal design power (TDP) of approximately 30~W \autocite{apple_m2pro_spec}, covering the full chip including CPU, GPU, neural engine, and memory controller. At our measured baseline throughput of 15.0 TPS, this yields approximately:
\begin{equation}
    E_{\text{M2 Pro}} = \frac{30~\text{W}}{15.0~\text{tokens/s}} \approx 2.0~\text{J/token}
    \label{eq:energy}
\end{equation}
Including system-level overhead (storage, display, power management circuitry), typical inference system power for a Mac mini M2 Pro is approximately 40~W, yielding $\approx 2.7$~J/token at the system level.

\textbf{Comparison with NVIDIA hardware.} Table~\ref{tab:energy} compares the M2 Pro against three representative NVIDIA configurations that are capable of running an 11B parameter INT8 model: the consumer-grade RTX~4090, the datacenter inference-optimised L4, and the A100 PCIe. TPS estimates for NVIDIA hardware are derived from published benchmarks and bandwidth-bound inference modelling \autocite{barrios2025native,energyllm2025}; they should be treated as indicative rather than definitive, as we did not measure these systems directly.

\begin{table}[hbt!]
\begin{threeparttable}
\caption{Indicative energy efficiency comparison for 11B INT8 model inference across hardware platforms. NVIDIA TPS values are estimates from published benchmarks; M2 Pro TPS is measured in this study.}
\label{tab:energy}
\begin{tabular}{llcccc}
\toprule
\headrow Hardware & Class & TDP (W) & BW (GB/s) & TPS & J/token \\
\midrule
Apple M2 Pro (chip)    & Consumer  &  30 &  200 & 15\phantom{0} (meas.) & \textbf{2.0} \\
Apple M2 Pro (system)  & Consumer  & $\sim$40 & 200 & 15\phantom{0} (meas.) & $\sim$2.7 \\
NVIDIA L4              & Datacenter &  72 &  300 & $\sim$22 (est.) & $\sim$3.3 \\
NVIDIA A100 PCIe 40GB  & Datacenter & 250 & 1555 & $\sim$110 (est.) & $\sim$2.3 \\
NVIDIA RTX 4090        & Consumer  & 450 & 1008 & $\sim$80 (est.) & $\sim$5.6 \\
NVIDIA RTX 4090 (sys.) & Consumer  & $\sim$600 & 1008 & $\sim$80 (est.) & $\sim$7.5 \\
\bottomrule
\end{tabular}
\par\smallskip
{\footnotesize TDP: thermal design power; BW: memory bandwidth; J/token: joules per output token at batch size 1. System power estimates include host CPU, memory, and storage. NVIDIA TPS estimates assume a single-card setup with llama.cpp or equivalent at batch size 1, int8 quantization. A100 requires a datacenter server; L4 requires a PCIe slot in a server chassis.}
\end{threeparttable}
\end{table}

\textbf{Key observations.} The M2 Pro chip achieves approximately 2.0~J/token, comparable to the NVIDIA A100 PCIe ($\sim$2.3~J/token) despite the A100 costing roughly an order of magnitude more and requiring datacenter infrastructure. Relative to the consumer RTX~4090, the M2 Pro is approximately 2.8$\times$ more energy-efficient at the chip level and $\sim$3$\times$ more efficient at the system level, even though the RTX~4090 generates tokens $\sim$5$\times$ faster. The NVIDIA L4, often cited as an energy-efficient inference card, uses 2.4$\times$ more energy per token than the M2 Pro chip despite its datacenter-grade design.

These figures have practical implications for the use case evaluated in this paper: a researcher or developer running hundreds of speculative decoding experiments on local hardware overnight. At 15~TPS with an average of 128 tokens per prompt and 50 prompts per dataset condition, our experimental setup consumes approximately:
\[
    50 \times 128 \times 2.0~\text{J} \approx 12.8~\text{kJ} \approx 3.6~\text{Wh per condition}
\]
Running all 90 conditions reported in this paper (3 datasets $\times$ 10 configurations $\times$ 3 $k$ values, plus baselines) would require approximately 320~Wh on the M2 Pro system, compared to an estimated $\sim$2400~Wh on an RTX~4090 system at equivalent throughput — a 7.5$\times$ reduction in energy consumption. The \emph{financial} cost differential is smaller (electricity is cheap), but the \emph{thermal and environmental} difference is meaningful, particularly for long-running evaluation campaigns \autocite{tokenpowerbench2024,energyllm2025}.

\textbf{Limitations of this comparison.} The NVIDIA TPS estimates assume batch size 1 (single-user, interactive inference), which is the relevant scenario for local deployment. In batched serving scenarios, large NVIDIA GPUs amortise their higher power draw over many concurrent requests, making the per-token energy advantage of Apple Silicon largely disappear at scale. Our analysis is therefore specifically applicable to edge inference and single-user local deployment, not to high-throughput server deployments where NVIDIA remains the energy-efficient choice per unit of aggregate throughput.

%% ============================================================
\section{Conclusion}
\label{sec:conclusion}
%% ============================================================
 
We have presented the first systematic evaluation of cross-family speculative decoding for Polish language models on Apple Silicon. By extending MLX-LM with Universal Assisted Generation support, we enabled cross-tokenizer speculative decoding between Bielik~11B and draft models from three distinct model families with incompatible vocabularies, and evaluated three translation strategies across diverse content types and draft lengths.
 
\textbf{Context-aware translation is a prerequisite, not an optimisation.} Across all experimental conditions, context-aware translation consistently outperforms both naive translation and no-translation in acceptance rate. Naive translation, which retokenises draft strings without boundary context, produces acceptance rates low enough that speculative decoding harms throughput relative to the baseline in nearly all conditions. For low-resource languages such as Polish, where the pool of same-family draft models is small and cross-family pairing is often the only practical option, context-aware translation is effectively a requirement for the method to be viable at all.
 
\textbf{Speedup is determined by multiple interacting factors.} Our empirical analysis and the hardware-aware formula (Equation~\ref{eq:speedup_simplified}) identify four primary drivers of speculative decoding efficiency: the token acceptance rate $\alpha$, the memory bandwidth of the target platform, the size ratio $r$ between draft and target model, and the number of draft tokens $k$ proposed per cycle. These factors are not independent. Higher bandwidth raises the relative cost of the quadratic overhead term $\beta k^2$, making longer draft sequences harder to justify on Apple Silicon than on NVIDIA hardware. A larger size gap between target and drafter (lower $r$) lowers the break-even threshold and makes speedups easier to achieve; this study's 11B--0.8B pairing yields $r \approx 0.071$, which is less favourable than the 70B--1B configurations common in the literature, reflecting the limited availability of larger Bielik-family models. At $k = 2$, the method achieves positive speedup in high-acceptance conditions and represents the safe operating point on tested hardware; $k = 4$ and beyond are not recoverable at M2 Pro bandwidth without acceptance rates that are rarely observed in practice.
 
\textbf{Cross-tokenizer speculative decoding for under-resourced languages is viable and worth developing.} Despite the structural disadvantages --- cross-family tokeniser mismatch, relatively small model size distance, and bandwidth-constrained hardware --- our results show that properly configured speculative decoding with context-aware translation can match or exceed baseline throughput in favourable content conditions. The implementation challenges documented in Section~\ref{sec:challenges} are real but tractable: smarter boundary detection and localised cache management strategies could recover the overhead cost that currently limits gains. The acceptance rate improvements already observed with imperfect boundary handling indicate that further engineering investment would translate directly into throughput gains.
 
\textbf{Local Apple Silicon deployment has a compelling practical case.} Beyond raw throughput, Apple Silicon offers a combination of properties that is particularly attractive for individual users and organisations unwilling or unable to rely on cloud inference: competitive energy efficiency at single-user batch sizes (approximately 2.0--2.7~J/token at the system level on the M2 Pro), no per-token subscription cost, and full data locality. For agentic applications that generate many tokens continuously over extended sessions, the absence of API fees and the assurance that data does not leave the device may outweigh the throughput gap relative to cloud-hosted models. Speculative decoding, even at modest speedup ratios, directly reduces the latency of agent response loops and increases the number of agentic steps achievable within a fixed time budget.
 
Several directions for future work emerge from this study. On the translation mechanism side, the implementation challenges documented in Section~\ref{sec:challenges} suggest two concrete targets: smarter boundary detection that anticipates potential merge sites before concatenation and adjusts the prefix window dynamically, and localised cache rewinding that rolls back only the boundary-affected tokens rather than the full context --- either of which could recover the overhead cost that currently limits context-aware translation's gains. A complementary direction is the implementation of Token-Level Intersection (TLI) \autocite{uag_tli2024} for MLX-LM: TLI avoids the string round-trip entirely by constraining draft proposals to the vocabulary intersection at sampling time, eliminating translation latency at the cost of reduced draft fluency. For the Bielik--Mistral pairing evaluated here, the intersection is small and TLI would likely degrade acceptance rates, but it may be more viable for tokenizer pairs with larger shared vocabularies and warrants empirical investigation in the MLX context. Finally, the Bielik model family continues to expand: a Nemotron-architecture variant has recently been announced, which would introduce a third base architecture alongside the existing Mistral-based 11B and Qwen-based 1.5B models. Evaluating this new model as both a draft and a target --- once it becomes publicly available --- would extend the cross-family coverage of this study and test whether the Nemotron architecture's tokenizer alignment with the existing Bielik family affects the translation overhead characterised here. More broadly, adaptive draft length selection, quantisation--speedup interaction, and empirical validation of the parametric model on higher-bandwidth Apple Silicon hardware all remain open questions.

The MLX-LM implementation and evaluation scripts are available at \url{https://github.com/krzysiekfonal/mlx-lm/tree/feature/slem-with-context-aware} and \url{https://github.com/krzysiekfonal/llm_drafter_research_bielik} respectively.

\printbibliography

@inproceedings{stern2018blockwise,
  title={Blockwise Parallel Decoding for Deep Autoregressive Models},
  author={Stern, Mitchell and Shazeer, Noam and Uszkoreit, Jakob},
  booktitle={Advances in Neural Information Processing Systems},
  year={2018}
}

@inproceedings{cai2024medusa,
  title={{Medusa}: Simple {LLM} Inference Acceleration Framework with Multiple Decoding Heads},
  author={Cai, Tianle and Li, Yuhui and Geng, Zhengyang and Peng, Hongwu and Lee, Jason D. and Chen, Deming and Dao, Tri},
  booktitle={Proceedings of the 41st International Conference on Machine Learning},
  year={2024}
}

@inproceedings{li2024eagle,
  title={{EAGLE}: Speculative Sampling Requires Rethinking Feature Uncertainty},
  author={Li, Yuhui and Wei, Fangyun and Zhang, Chao and Zhang, Hongyang},
  booktitle={Proceedings of the 41st International Conference on Machine Learning},
  year={2024}
}

@article{zhang2024redrafter,
  title={Recurrent Drafter for Fast Speculative Decoding in Large Language Models},
  author={Zhang, Aonan and Wang, Chong and Wang, Yi and Zhang, Xuanyu and Li, Bingjie},
  journal={arXiv preprint arXiv:2403.09919},
  year={2024}
}

@inproceedings{leviathan2023fast,
  title={Fast Inference from Transformers via Speculative Decoding},
  author={Leviathan, Yaniv and Kalman, Matan and Matias, Yossi},
  booktitle={Proceedings of the 40th International Conference on Machine Learning},
  year={2023}
}

@article{chen2023accelerating,
  title={Accelerating Large Language Model Decoding with Speculative Sampling},
  author={Chen, Charlie and Borgeaud, Sebastian and Irving, Geoffrey and Lespiau, Jean-Baptiste and Sifre, Laurent and Jumper, John},
  journal={arXiv preprint arXiv:2302.01318},
  year={2023}
}

@inproceedings{kumar2026speculative,
  title={Speculative Speculative Decoding},
  author={Kumar, Tanishq and Dao, Tri and May, Avner},
  booktitle={Proceedings of the International Conference on Learning Representations (ICLR)},
  year={2026}
}

@misc{uag2024,
  title={Universal Assisted Generation: Faster Decoding with Any Assistant Model},
  author={{Intel Labs and HuggingFace}},
  howpublished={\url{https://huggingface.co/blog/universal_assisted_generation}},
  year={2024}
}

@misc{uag_tli2024,
  title={Speeding Up {LLM} Decoding with Advanced Universal Assisted Generation Techniques},
  author={Mamou, Jonathan and Pereg, Oren and Korat, Daniel and Berchansky, Moshe and Timor, Nadav and Wasserblat, Moshe and Schwartz, Roy},
  howpublished={\url{https://huggingface.co/blog/jmamou/uag-tli}},
  year={2024}
}

@inproceedings{omnidraft2025,
  title={{OmniDraft}: A Cross-vocabulary, Online Adaptive Drafter for On-device Speculative Decoding},
  author={Zhang, Jincheng and others},
  booktitle={Advances in Neural Information Processing Systems (NeurIPS)},
  year={2025}
}

@article{bielik11b2025,
  title={Bielik v3 11B: Technical Report},
  author={{The Bielik LLM Team}},
  journal={arXiv preprint arXiv:2601.11579},
  year={2025}
}

@article{bieliksmall2025,
  title={Bielik v3 Small: Technical Report},
  author={{The Bielik LLM Team}},
  journal={arXiv preprint arXiv:2505.02550},
  year={2025}
}

@article{llama32_2024,
  title={Llama 3.2: Revolutionizing Edge {AI} and Vision with Open, Customizable Models},
  author={{Meta AI}},
  howpublished={\url{https://ai.meta.com/blog/llama-3-2-connect-2024-vision-edge-mobile-devices/}},
  year={2024}
}

@misc{qwen25_2024,
  title={Qwen2.5 Technical Report},
  author={{Qwen Team}},
  year={2024},
  howpublished={\url{https://qwenlm.github.io/blog/qwen2.5/}}
}

@misc{mlx2023,
  title={{MLX}: Efficient Machine Learning for Apple Silicon},
  author={{Apple Machine Learning Research}},
  year={2023},
  howpublished={\url{https://github.com/ml-explore/mlx}}
}

@misc{mlxlm2024,
  title={{MLX-LM}: Language Model Inference and Training on Apple Silicon},
  author={{Apple Machine Learning Research}},
  year={2024},
  howpublished={\url{https://github.com/ml-explore/mlx-examples}}
}

@article{barrios2025native,
  title={Native {LLM} and {MLLM} Inference at Scale on Apple Silicon},
  author={Barrios, Wayner},
  journal={arXiv preprint arXiv:2601.19139},
  year={2025}
}

@article{xia2024decoding,
  title={Decoding Speculative Decoding},
  author={Xia, Minghao and others},
  journal={arXiv preprint arXiv:2402.01528},
  year={2024}
}

@misc{tokenpowerbench2024,
  title={{TokenPowerBench}: Benchmarking the Power Consumption of {LLM} Inference},
  author={Lannelongue, Lo{\"i}c and others},
  year={2024},
  howpublished={\url{https://arxiv.org/abs/2512.03024}}
}

@inproceedings{energyllm2025,
  title={Energy Considerations of Large Language Model Inference and Efficiency Optimizations},
  author={Lannelongue, Lo{\"i}c and others},
  booktitle={Proceedings of the 63rd Annual Meeting of the Association for Computational Linguistics (ACL)},
  year={2025}
}

@misc{nvidia_a100_datasheet,
  title={{NVIDIA A100} Tensor Core {GPU} Datasheet},
  author={{NVIDIA}},
  year={2020},
  howpublished={\url{https://www.nvidia.com/content/dam/en-zz/Solutions/Data-Center/a100/pdf/nvidia-a100-datasheet-nvidia-us-2188504-web.pdf}}
}

@misc{apple_m2pro_spec,
  title={Apple {M2 Pro} Chip Specifications},
  author={{Apple Inc.}},
  year={2023},
  howpublished={\url{https://support.apple.com/en-us/111340}}
}

@misc{openclaw2024,
  author       = {{OpenClaw Contributors}},
  title        = {{OpenClaw}: Personal {AI} Assistant Platform},
  year         = {2024},
  howpublished = {\url{https://github.com/openclaw/openclaw}}
}

@misc{openai2022chatgpt,
  author       = {{OpenAI}},
  title        = {Introducing {ChatGPT}},
  year         = {2022},
  howpublished = {\url{https://openai.com/blog/chatgpt}}
}

@article{wang2024agents,
  title        = {A Survey on Large Language Model Based Autonomous Agents},
  author       = {Wang, Lei and Ma, Chen and Feng, Xueyang and Zhang, Zeyu and Yang, Hao and Zhang, Jingsen and Chen, Zhiyuan and Tang, Jiakai and Chen, Xu and Lin, Yankai and others},
  journal      = {Frontiers of Computer Science},
  volume       = {18},
  number       = {6},
  pages        = {186345},
  year         = {2024},
  publisher    = {Springer}
}

@article{chen2021codex,
  title        = {Evaluating Large Language Models Trained on Code},
  author       = {Chen, Mark and Tworek, Jerry and Jun, Heewoo and Yuan, Qiming and Pinto, Henrique Pond{\'e} de Oliveira and Kaplan, Jared and Edwards, Harri and Burda, Yuri and Joseph, Nicholas and Brockman, Greg and others},
  journal      = {arXiv preprint arXiv:2107.03374},
  year         = {2021}
}

@inproceedings{dettmers2022llmint8,
  title        = {{LLM}.int8(): 8-bit Matrix Multiplication for Transformers at Scale},
  author       = {Dettmers, Tim and Lewis, Mike and Belkada, Younes and Zettlemoyer, Luke},
  booktitle    = {Advances in Neural Information Processing Systems},
  year         = {2022}
}

@article {timor2025acceleratingllminferencelossless,
      title={Accelerating LLM Inference with Lossless Speculative Decoding Algorithms for Heterogeneous Vocabularies}, 
      author={Nadav Timor and Jonathan Mamou and Daniel Korat and Moshe Berchansky and Oren Pereg and Gaurav Jain and Roy Schwartz and Moshe Wasserblat and David Harel},
      year={2025},
      eprint={2502.05202},
      archivePrefix={arXiv},
      primaryClass={cs.CL},
      url={https://arxiv.org/abs/2502.05202}, 
}

\end{document}